\documentclass[runningheads]{llncs}
\usepackage[T1]{fontenc}
\usepackage{graphicx}
\usepackage{cite}
\usepackage{amsmath}

\usepackage{amssymb,amsfonts}
\usepackage{algorithmic}
\usepackage{textcomp}
\usepackage{xcolor}

\usepackage[hyphenbreaks]{breakurl}
\usepackage[hyphens]{url}
\usepackage{orcidlink}
\begin{document}

\title{CURVETE: Curriculum Learning and Progressive Self-supervised Training for Medical Image Classification}
\titlerunning{CURVETE: Curriculum Learning and Progressive Self-supervised Training}
%
%
%
\author{Asmaa Abbas\orcidlink{0000-0002-7792-3976}\inst{1} \and
Mohamed Medhat Gaber\orcidlink{0000-0003-0339-4474}\inst{1} \and
Mohammed M. Abdelsamea\orcidlink{0000-0002-2728-1127}\inst{2}}


\authorrunning{Abbas et al.}

\institute{Birmingham City University, Birmingham, UK\\
\email{\{asmaa.husien,mohamed.gaber\}@bcu.ac.uk} \and
University of Exeter, Exeter, UK\\
\email{m.abdelsamea@exeter.ac.uk}}
\maketitle              
\begin{abstract}
Identifying high-quality and easily accessible annotated samples poses a notable challenge in medical image analysis. Transfer learning techniques, leveraging pre-training data, offer a flexible solution to this issue. However, the impact of fine-tuning diminishes when the dataset exhibits an irregular distribution between classes. This paper introduces a novel deep convolutional neural network, named Curriculum Learning and Progressive Self-supervised Training (\emph{CURVETE}). \emph{CURVETE} addresses challenges related to limited samples, enhances model generalisability, and improves overall classification performance. It achieves this by employing a curriculum learning strategy based on the granularity of sample decomposition during the training of generic unlabelled samples. Moreover, \emph{CURVETE} address the challenge of irregular class distribution by incorporating a class decomposition approach in the downstream task. The proposed method undergoes evaluation on three distinct medical image datasets: brain tumour, digital knee x-ray, and Mini-DDSM datasets. We investigate the classification performance using a generic self-supervised sample decomposition approach with and without the curriculum learning component in training the pretext task. Experimental results demonstrate that the \emph{CURVETE} model achieves superior performance on test sets with an accuracy of 96.60\% on the brain tumour dataset, 75.60\% on the digital knee x-ray dataset, and 93.35\% on the Mini-DDSM dataset using the baseline ResNet-50. Furthermore, with the baseline DenseNet-121, it achieved accuracies of 95.77\%, 80.36\%, and 93.22\% on the brain tumour, digital knee x-ray, and Mini-DDSM datasets, respectively, outperforming other training strategies.

\keywords{Curriculum learning \and Convolutional neural networks \and Data irregularities \and Medical image classification \and Self-supervision learning.}
\end{abstract}
\section{Introduction}
Medical image classification using deep learning plays a critical role in the healthcare domain for identifying and diagnosing various diseases and conditions such as tumours, fractures, and abnormalities. Brain tumours are one of the most difficult diseases to treat, and they can have a potential impact on essential functions due to the complex nature of the brain. Furthermore, early detection of knee diseases is important to prevent further damage and complications. Moreover, breast cancer is extremely common in many countries, particularly among women. Based on that, several research works have been introduced in medical imaging for the early detection of diseases. Deep learning algorithms, especially convolutional neural networks (CNNS), have significantly revolutionised various medical image classification tasks. CNNs can automatically learn hierarchical representations and complex features from the dataset without the need for explicit feature engineering. The most effective way to train a CNN architecture is to transfer the knowledge gained from a previously trained network to a new task, especially when dealing with small amounts of labelled data, such as medical images. However, when the source and target domains are not related, the pre-trained features may not be informative for the target task.

Self-supervised learning (SSL) has gained significant attention in various computer vision tasks, including image classification, especially when obtaining labelled data is expensive or time-consuming \cite{liu2021self}. It can effectively transfer its meaningful representations or features from unlabelled data (pretext task) to a new task with fewer labelled examples (downstream task). However, when one or more classes have significantly fewer samples than others, it can be difficult to build a robust classification model for datasets due to potential biases in model training and evaluation. Class decomposition approaches aim to address issues associated with unequal class distributions within a dataset by improving boundary learning between specified classes and making sure that machine learning models can effectively learn and understand all local patterns within each class in the dataset \cite{abbas2020detrac}. The decomposition mechanism works by breaking down the original classes in a dataset into simpler sub-classes. Then, each sub-class is assigned a new label linked to its original class and treated as a separate new class. Following training, those sub-classes are recollected to compute the final prediction's error correction.

Curriculum learning (CL) is a strategy for training a machine learning model that can be used to speed up the learning process and improve the generalisation performance of the model by introducing the training samples in a meaningful order rather than a random or fixed order \cite{elman1993learning, graves2017automated}. The motivation behind this is derived from how humans learn, which frequently begins with simpler concepts before progressing to more complex ones \cite{bengio2009curriculum}. Likewise, in traditional CL, simple pattern examples are introduced first during training, followed by more complex and challenging ones. In contrast, the anti-CL strategy reverses this approach, starting with difficult examples and gradually progressing to simpler ones \cite{mermer2017scalable}. This reverse technique encourages the model to learn robust features and representations early in training, potentially leading to faster convergence. 

Based on this concept, we previously introduced \emph{CLOG-CD}, in which a pre-trained backbone was used to initialise the network, and an anti-CL strategy with class decomposition was applied in the downstream task \cite{abbas2025clog}. \emph{CLOG-CD} was the first to leverage anti-CL to guide class decomposition by gradually increasing class complexity in a structured way. Where the model starts training at the hard level (maximum granularity level), and the learnt weights are then gradually transformed to the next lower granularity level until reaching the easiest level (lowest-level granularity) and then return to the maximum granularity level. This progressive manner allows the model to understand relationships between examples and reduces the impact of overlapping class distributions, hence improving the generalisation on unseen data. 

In this work, we adopt a similar strategy to propose a novel training approach, called \emph{CURVETE}: Curriculum Learning and Progressive Self-Supervised Training. Unlike \emph{CLOG-CD}, which relied on a pre-trained backbone, \emph{CURVETE} utilises an SSL pipeline by integrating sample decomposition and anti-CL with different levels of granularity to train a large set of unlabelled samples. This mechanism enables the model to improve the feature representations and build stronger prior knowledge before fine-tuning to another small dataset. In addition, by applying the anti-CL strategy and adjusting granularity levels, the optimiser can effectively explore a broader solution space. This facilitates the discovery of new patterns and allows for more effective fine-tuning of the model weights. In simple words, at higher granularity levels, the model focuses on specific features by learning to classify smaller sub-classes, while at lower granularity levels, more generic features are analysed. This approach empowers the model to adapt to complex patterns that may have been difficult to extract in traditional deep neural network learning. Finally, the learnt information from the pretext task is utilised to solve a downstream task with limited labelled data. In the downstream task, we also incorporated the anti-CL strategy with the decomposition method to simplify the complex structure of the dataset, effectively reducing overlap in class distributions and improving classification with limited labelled data. The code of this work is available at (\url{https://github.com/ascodeuser/CURVETE}). Our contributions are summarised as follows:
\begin{itemize}

    \item \emph{CURVETE} involves utilising the anti-CL strategy in training the pretext model, which encourages the training process to be more effective in learning rich and meaningful representations, leading to faster convergence and improved performance in the downstream task.

    \item \emph{CURVETE} introduces the samples in a progressive manner, from hard-to-easy order based on a gradual decomposition approach, which helps the model to better understand the class boundaries between classes before handling more complex patterns.
    
   \item \emph{CURVETE} can handle irregularities in data distribution by adapting the granularity of class decomposition, resulting in improved model performance.

    \item The extensive experiments based on three different medical image datasets using two different baselines provide a promising and generalised solution by transferring knowledge from generic image recognition tasks to more specific problems.
 
\end{itemize}

The rest of this paper is organised as follows: Section 2 presents previous works of the SSL method for medical image classification. Section 3 covers the main components of the proposed method. Section 4 describes the results of the experiments based on three different medical image datasets. Section 5 discusses and concludes our work.

\section{Related Work}

SSL has achieved great success in many works, where the network learns meaningful information from the pretext training task that can be useful later in different computer vision tasks \cite{jing2020self, xu2021review}. For instance, in \cite{koohbanani2021self}, a self-path model was introduced to enhance the classification performance on a small number of pathology images. In \cite{mishra2023ssclnet}, the authors used contrastive discriminative methods to enable the pretext model to extract meaningful features from the latent space, and then fine-tune the learnt knowledge on a small brain tumour dataset. In addition, in \cite{abbas20214s} we previously proposed a self-supervised model that used the sample decomposition method to generate pseudo-labels from large chest x-ray images. We used ResNet18 to train the pretext model and then transferred the learnt features to a small dataset where limited annotated COVID-19 was available. 

Another remarkable method for medical image classification tasks is the CL strategy. For example, in \cite{lotter2017multi}, two models were built to classify breast cancer. The first model was used to extract the features based on the image patches and then transfer those features into another model for the classification task. In \cite{jesson2017cased}, a (CASED) model was introduced to detect pulmonary nodules in a CT image dataset. The CL strategy used the complexity of the input nodules, where the model learnt to recognise nodules from their surroundings and then gradually introduced a more global context. In \cite {luo2022deep}, CL was applied based on the difficulty of the classification task, where the model trained first from the easy task (binary classification) to the hard task (multi-classification) using a pre-trained VGG-19 network. In \cite{tang2018attention}, the method was designed around the severity of diseases, beginning with severe cases, followed by moderate and mild ones, thereby utilising prior knowledge to guide the training process. In \cite{park2019curriculum}, CL was applied based on the classification task from easy to hard to improve the detection of pulmonary abnormalities from chest x-ray images. They first trained on patch images around regions of interest to focus on thoracic abnormalities, then fine-tuned the model with full images. In \cite{srinidhi2021improving}, the authors introduced (HaDCL) to enhance histology image classification performance. They used two different SSL techniques for training the unlabelled set using ResNet-18, before fine-tuning these learned representations on the downstream task. The CL was applied on the labelled dataset, where the difficulty of training samples was determined based on their loss values from easy to hard and from hard to very hard samples. 

\section{Methodology}\label{sec:methods1}
\begin{figure*}[h!]
  \centering
     \includegraphics[scale=0.51]{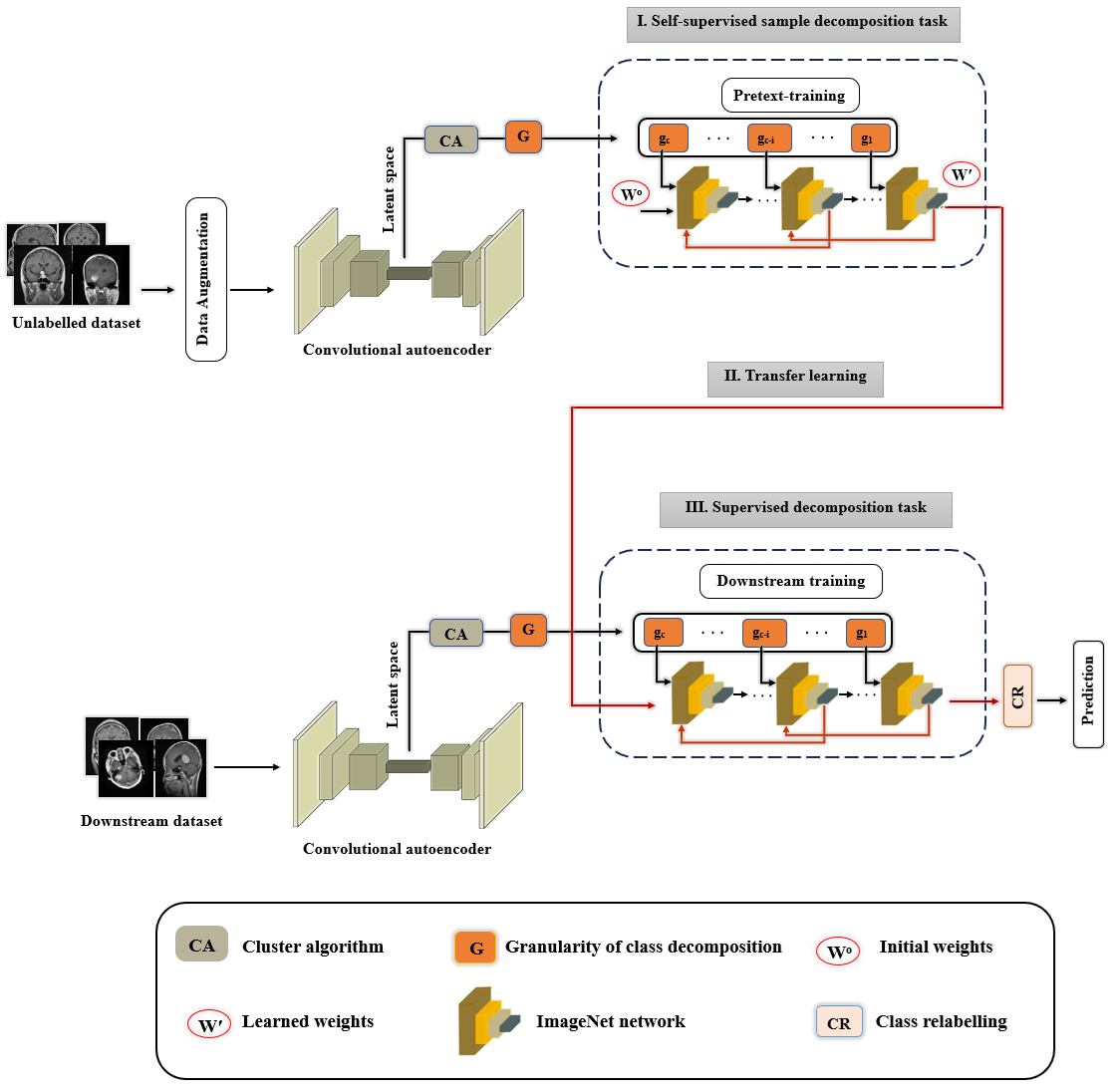} 
    \caption{The framework of the \emph{CURVETE} model, where $g_{c}$ refers to the maximum number of decomposition granularities of the classes.}
    \label{Method}
\end{figure*}

This section represents our proposed method \emph{CURVETE} in detail and the performance estimation metrics that we used to evaluate our model; see Fig. \ref{Method}.

\subsection{Self-Supervised Pretext Task Learning}

Our proposed method starts by extracting local feature representations from a large collection of unlabelled medical images using a convolutional autoencoder (CAE) model. The CAE effectively compresses high-dimensional input images into lower-dimensional representations, highlighting critical features while preserving essential structural details. Moreover, its reconstruction capability ensures robustness against variations in object position or orientation, making it suitable for handling irregularities in medical image datasets.

The extracted features are subsequently used to generate pseudo-labels through granularity-based sample decomposition. To perform this decomposition, we employed the $k$-means clustering algorithm, which partitions the feature space into $k$ clusters \cite{wu2008top}. Each data point ($x_{i}$) is assigned to the cluster with the closest centroid ($c$) based on minimising the squared Euclidean distance (SED).

\begin{equation}
     SED= \sum _{j=1}^{k} \sum _{i=1}^{n}\parallel x_{i}^{ \left( j \right) }-c_{j}\parallel^{2}, 
    \end{equation}

\begin{figure*} [h!]
\centering
     \centerline{\includegraphics[scale=0.45]{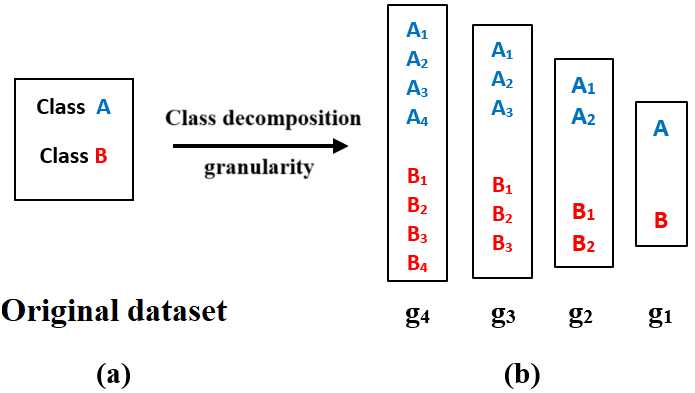}}

    \caption{An example illustrates the granularity of the class decomposition process: a) the original dataset with two classes, $A$ and $B$; b) the new datasets generated after applying class decomposition granularity for $k$=4. At $g_{4}$, each class is divided into four sub-classes. Similarly, the granularity at each subsequent level $g_{3}$ and $g_{2}$, and finally $g_{1}$ corresponds to the original classes without any decomposition.} 
    \label{granularity}
\end{figure*}

The granularity of decomposition is controlled by the parameter $k$, which determines the number of clusters in each new decomposed dataset. Let $\textbf{G}$ be the granularity sequence derived from the latent space features extracted by the CAE. 
We aim to decompose it into multiple levels, defined as: $\{\text{k}, \text{k-1}, \text{k-2}, \dots, \text{1} \}$, arranged in descending order. Each level of $G$ corresponds to a dataset decomposed at a specific granularity level. For example, if $k=4$, the granularity sequence produces three additional datasets $\{\text{g}_4, \text{g}_3, \text{g}_2\}$ along with the original dataset $g_{1}$. Here, $g_{4}$ represents the dataset with the highest level of decomposition (i.e, each class split into four sub-classes), while $g_{1}$ corresponds to the original dataset without decomposition, see Fig. \ref{granularity}. The generated pseudo-labels from this process provide a structured progression of training data, forming the basis for the self-supervised pretext task.

For the unlabelled datasets, we investigated the performance using values of $k$=5 and $k$=10, leading to decomposition levels with five and ten granularity levels, respectively, along with the original dataset ($g_{1}$). Additionally, to classify the pseudo-labels and train the pretext task, we employed two distinct baseline networks: ResNet-50 \cite{he2016deep} and DenseNet-121 \cite{huang2017densely}, which serve as the backbone to facilitate coarse transfer learning.

\subsection{Curriculum Learning:}


In the context of CL, there are two main factors involved: the scoring and pacing functions. a) The ``scoring function'', also known as the training scheduler, evaluates the difficulty of examples in the dataset and guides the order in which they are presented to the model during training. b) The ``pacing function'' controls the rate at which more difficult examples are introduced to the model as it progresses through training.

The scoring function in the case of descending-ascending order can be written as: 
\begin{equation}  
     d(x_{i},y_{i}) > d(x_{j},y_{j}),   \forall \, S: X \to R,
\end{equation}

where the data point $d(x_{i},y_{i})$ is considered more challenging than $d(x_{j},y_{j})$. 
During the training process, we employed the mini-batch stochastic gradient descent (mSGD) method, where the model parameters were updated using small batches of data at each iteration. Consequently, the pacing function (P) for the subset of samples $X\acute{_{i}}$ within the mini-batches (MB) can be defined as:
  
\begin{equation}
    P_{\theta}(i)= \left| X\acute{_{i}} \right|
\end{equation}

where $\theta$ is the trainable parameters, and ${X\acute{_{i}}=\{X\acute{_{1}},X\acute{_{2}},...,X\acute{_{M}}}\}$ includes the samples from batch $B_{i}$ sorted by the scoring function in descending order.

Following the results of \emph{CLOG-CD}, where we investigated performance with different learning speeds, the experiments showed that the single-speed step consistently achieved the highest performance on different datasets. Based on these findings, \emph{CURVETE} adopts the single-speed strategy, ensuring a more stable and effective learning process. The training process begins at the highest granularity ($g_{k}$), where the unlabelled samples are divided into the maximum number of clusters. At this stage, the model focuses on specific features by learning through smaller sub-classes. The learnt knowledge is then gradually transferred to lower granularity levels (e.g., $g_{k-1}$), refining the features at each step until reaching the final level, where more general patterns are analysed.


\subsection{Downstream Training}

The downstream dataset benefits from the learnt knowledge obtained during the pretext model training, enabling better predictions with fewer labelled examples. Here, \emph{CURVETE} also incorporates the anti-CL strategy with the class decomposition method to train on smaller subsets of downstream data. To generate different granularities of decomposition, we also extracted feature representations from the latent space of the CAE and applied the $k$-means algorithm with component ($k$=5). The class decomposition method first helps the model learn specific features by simplifying the complex structure of the dataset and defining clear boundaries between classes. This makes it easier for the model to understand relationships between examples and reduces the impact of overlapping class distributions. Finally, class relabelling is performed to correct the classification predictions made during the decomposition process, and ensure that the final output corresponds to the initial classification problem.
For training the downstream task, the training process was repeated over 20 times for the baseline ResNet-50 and 10 times for the baseline DenseNet-121.

\section{Experimental Details and Results}\label{results}

This section describes the datasets used to evaluate the effectiveness of our method, \emph{CURVETE}, which utilises anti-CL with sample decomposition for training the pretext task. In addition, it discusses the experimental results obtained from the test sets of three different medical image datasets using two different granularity components. Finally, we compare our findings against other training strategies, including ablation studies and state-of-the-art methods.


\subsection{Datasets Description}
In this work, we evaluated our proposed method on three different datasets: brain tumour, knee x-ray images, and digital mammogram image datasets.

\begin{itemize}

\item \textbf{Brain tumour images dataset}, we used a public brain tumour dataset as unlabelled samples available at:(\url{https://www.kaggle.com/datasets/navoneel/brain-mri-images-for-brain-tumor-detection}). More samples were generated by applying data-augmentation (AUG) processes such as rotation, transformation, reflection, and sharpening. This technique produced 59,229 images of brain tumours. For the labelled dataset, we used the dataset from Nanfang and General Hospitals, Tianjin Medical University, China, with three types of acquired brain tumours \cite{badvza2020classification}, see Table \ref{distribution}. All the images are $400 \times 400$ pixels with PNG format. 

\item \textbf{knee x-ray images dataset}: we used the knee Osteoarthritis dataset with Severity Grading (OAI) as unlabelled samples \cite{chen2018knee}. The dataset contains a total of 9786 images categorised into five grades. We also generated more samples by applying several AUG techniques, such as reflection, shifting, sharpening, and rotation, to produce 68,502 samples of the knee x-ray dataset. For the labelled dataset, we used the digital knee x-ray images dataset from reputable hospitals and diagnostic centres \cite{gornale2020digital} using PROTEC PRS 500E x-ray machine with the help of two medical experts, 8-bit gray-scale graphics were used in the original images in PNG format. 
In our work, we used the images contained in the sub-folder “MedicalExpert-I” which are labelled into five classes, see Table \ref{distribution}.

\item \textbf{Digital mammograms dataset:} MIAS mammograms dataset, used as unlabelled samples \cite{suckling1994mammographic}, was augmented through processes such as cropping, zooming, reflection, shifting, and rotation to create 47,334 samples. For the labelled dataset, we used the Mini-DDSM dataset, which is a subset of the larger Digital Database for Screening Mammography (DDSM) \cite{lekamlage2020mini}. The dataset is divided into three classes: Normal, Cancer, and Benign, see Table \ref{distribution}, and all images come in JPEG format with dimensions between 125 and 320 pixels. 
\end{itemize}




For clear visibility, we used histogram equalisation to adjust the distribution and enhance the contrast. Each downstream dataset was randomly split into 70\% and 20\% for training and validation sets, respectively, and 10\%  as a test set for performance evaluation.


\begin{table}[htbp]
\begin{center} 
\caption{The distribution of each labelled dataset.}
\label{distribution}
\renewcommand{\arraystretch}{1.0}
\begin{tabular}{ccccc}
\hline
Dataset  & Training  & Validation & Test & Overall\\
\hline
Brain tumour  & 1960  & 489 & 615   & 3064 \\
Digital knee x-ray &  1188 & 294 & 168& 1,650  \\
Mini-DDSM & 5622  & 1404  & 782 & 7,808 \\
\hline
\end{tabular}
\end{center} 
\end{table}

\subsection{Hyper-Parameter Settings}

We used a CAE model with two convolutional layers to extract deep local features from unlabelled samples and generate pseudo-labels for pretext training. For the brain tumour and Mini-DDSM datasets, the first and second layers had 16 and 8 filters, respectively, while for the knee x-ray dataset, the layers had 32 and 16 filters. Models were trained with a $3\times3$ kernel for 50 epochs using the Adam optimiser with ReLU activation and a learning rate of 0.001. The features from the latent space were then clustered using $k$-means to generate pseudo-labels for classification. For unlabelled samples, we trained the model using two values of $k$=5 and $k$=10, utilising ResNet-50 and DenseNet-121 as backbone architectures with their pre-trained weights. The pretext model was trained using an anti-CL strategy with sample decomposition, starting at the highest granularity level and progressively transitioning to lower levels until reaching the original classes. This process was repeated for 10 iterations in both directions. Please note that the selected values of $k$ (i.e., 5 and 10) were arbitrarily chosen to control the granularity decomposition sequence and explore diverse patterns across different levels of granularity.

For the labelled datasets, we used the same CAE models to extract the local features, then applied $k$-means clustering ($k=5$) to generate different levels of granularity. The anti-CL strategy was also used in training the downstream task, starting with the maximum granularity $(g_{5})$. Then, the learnt weights are gradually transferred until reaching the original classes $(g_{1})$. This process was repeated 20 times for ResNet-50 and 10 times for DenseNet-121.
Based on trial-and-error experiments, the learning rate for training the brain tumour dataset was set to 0.001, with a weight decay of 0.85 applied every 15 epochs. For the digital knee x-ray images, the learning rate was set to 0.01, with a weight decay of 0.90 applied every 15 epochs. Finally, for the Mini-DDSM dataset, the model was trained with a learning rate of 0.001 and a weight decay of 0.90 applied every 15 epochs.

\subsection{Performance Evaluation}
To assess the effectiveness of our proposed method, we utilised various performance metrics, including accuracy, precision, recall, and F1 score for the multi-classification tasks \cite{sokolova2009systematic}. All experiments were conducted using the Python programming, Keras library with 50 mini-batch size, and 50 epochs with (mSGD) optimiser \cite{robbins1951stochastic} and cross-entropy loss function. 

\subsection{Performance of \emph{CURVETE} Model}



We evaluated the performance of \emph{CURVETE} on the downstream task after training the pretext model using two different components ($k$ = 5 and $k$ = 10). The results are summarised in Table \ref{results_Resnet} and Table \ref{res_DenseNet}. As shown in the tables, the values in bold indicate the highest performance scores achieved. For the brain tumour dataset, \emph{CURVETE} with ResNet-50 achieved significant performance with ACC, PR, RE, and F1 values of 96.60\%, 95.82\%, 96.56\%, and 96.19\%, respectively. Similarly, with DenseNet-121, the model achieved the highest performance at 95.77\%, 95.18\%, 95.15\%, and 95.16\% for ACC, PR, RE, and F1-score, respectively. 
For further investigation, we evaluated the performance of \emph{CURVETE} on the 168 test set of knee x-ray images. The best performance was observed with DenseNet-121 with 80.36\%, 83.24\%, 78.64\%, 80.87\% for ACC, PR, RE, and F1-score, respectively; see Table \ref{res_DenseNet}. For ResNet-50, the performance metrics recorded were 75.60\% for ACC, 76.54\% for PR, 73.54\% for RE, and 75.01\% for F1-score. 
Finally, we evaluated the performance of \emph{CURVETE} on the Mini-DDSM dataset, which consists of 786 test images. With ResNet-50, the model achieved an accuracy of 93.35\%, closely matching the 93.22\% accuracy obtained with DenseNet-121. 

\begin{table}[ht!] 
\begin{center} 
\caption{The classification performance of \emph{CURVETE} using the ResNet-50 baseline network on the downstream datasets.}
\label{results_Resnet} 
\small
\tabcolsep=0.10cm
\resizebox{8.8cm}{!}{
\renewcommand{\arraystretch}{1.3}
\begin{tabular}{c |  c c c c | c c c c} 
\hline 
Dataset  &  \multicolumn{8}{c}{\emph{CURVETE}} \\
&  \multicolumn{4}{c}{pesudo-labells ($k$=5)} & \multicolumn{4}{c}{pesudo-labells ($k$=10)}\\
& ACC & PR & RE & F1 & ACC & PR & RE & F1 \\
 &($\%$) & ($\%$) & ($\%$) & ($\%$) &  ($\%$) & ($\%$) & ($\%$) & ($\%$) \\
\hline 
brain tumour &  95.12 & 94.11 & 95.04  & 94.57 & \textbf{96.60} & \textbf{95.82} & \textbf{96.56} & \textbf{96.19} \\

digital knee x-ray &   \textbf{75.60}  & \textbf{76.54} & \textbf{73.54} & \textbf{75.01}  & 73.21& 75.06 & 73.07  & 74.05 \\

Mini-DDSM &  \textbf{93.35} & \textbf{93.35} & \textbf{93.55} & \textbf{93.45} &  91.94 & 92.04 & 92.12 & 92.08 \\

\hline 
\end{tabular}
}
\end{center} 
\end{table} 


\begin{table}[h!]  
\begin{center} 
\caption{The classification performance of \emph{CURVETE} using the DenseNet-121 baseline on the downstream datasets.}
\label{res_DenseNet} 
\small
\tabcolsep=0.10cm
\resizebox{8.8cm}{!}{
\renewcommand{\arraystretch}{1.3}
\begin{tabular}{c |  c c c c | c c c c} 
\hline 
Dataset  &  \multicolumn{8}{c}{\emph{CURVETE}} \\
&  \multicolumn{4}{c}{pesudo-labells ($k$=5)} & \multicolumn{4}{c}{pesudo-labells ($k$=10)}\\
& ACC & PR & RE & F1 & ACC & PR & RE & F1 \\
 &($\%$) & ($\%$) & ($\%$) & ($\%$) &  ($\%$) & ($\%$) & ($\%$) & ($\%$) \\
\hline 
brain tumour & \textbf{95.77} & \textbf{95.18} & \textbf{95.15} & \textbf{95.16} & 93.01  & 91.79 & 92.48 & 92.13     \\
digital knee x-ray & \textbf{80.36} & \textbf{83.24} & \textbf{78.64} & \textbf{80.87} & 72.62  & 71.04 &  68.14& 69.56      \\
Mini-DDSM & 92.58 & 92.63 & 92.79 & 92.71  & \textbf{93.22} & \textbf{93.29} & \textbf{93.40} & \textbf{93.35}    \\
\hline 
\end{tabular}
}
\end{center} 
\end{table} 

\subsection{Ablation Study}
In the ablation study, we compared our model with three different training strategies: (1) traditional transfer learning using a pre-trained network; (2) the \emph{CLOG-CD} model; and (3) training the pretext model based on sample decomposition without using the anti-CL strategy on the unlabelled dataset, which we referred to as \emph{CURVETE}(WO/CL, W/SD). Table \ref{Ablation_Resnet} and Table \ref{Ablation_DenseNet} demonstrate the performance of these models using ResNet-50 and DenseNet-121, respectively. As shown, the traditional transfer learning technique consistently achieved the lowest performance in all datasets, demonstrating its limitations in handling a small number of samples and irregular class distributions. However, the \emph{CLOG-CD} model performed better than the traditional transfer learning technique due to leveraging anti-CL and class decomposition at different levels of granularity in the downstream task. This progressive structure improves generalisation and the training process by gradually increasing class complexity in a structured way. Finally, \emph{CURVETE}(WO/CL, W/SD), shows notable improvement in the digital knee x-ray dataset and Mini-DDSM datasets. This confirms that utilising SSL with sample decomposition for training unlabelled data encourages the transformation of coarse features from general samples to specific tasks by simplifying the complex patterns and local structure of the dataset, providing more effective knowledge before fine-tuning for the subsequent task. 
By comparing these results with those obtained from \emph{CURVETE}, we observe that incorporating the anti-CL strategy into SSL with sample decomposition offers a promising solution for enhancing feature transferability from the pretext task and improving generalisation on new datasets, particularly in scenarios with limited and irregular data distributions.

\begin{table*}[htbp] 
\begin{center} 
\caption{The classification performance of other training strategies using the ResNet-50 baseline.}
\label{Ablation_Resnet} 
\small
\tabcolsep=0.10cm
\resizebox{12.0cm}{!}{
\renewcommand{\arraystretch}{1.1}
\begin{tabular}{c |  c  c  c c | c c c c | c c c c } 
\hline 

Dataset  & \multicolumn{4}{c|}{Traditional transfer learning } & \multicolumn{4}{c|}{\emph{CLOG-CD}} & \multicolumn{4}{c}{ \emph{CURVETE}(WO/CL, W/SD)}  \\
&  \multicolumn{4}{c|}{}  & \multicolumn{4}{c|}{} \\

& ACC & PR & RE & F1 & ACC & PR & RE & F1 & ACC & PR & RE & F1 \\
 &($\%$) & ($\%$) & ($\%$) & ($\%$) &  ($\%$) & ($\%$) & ($\%$) & ($\%$) & ($\%$) & ($\%$) & ($\%$) & ($\%$)  \\

\hline 
brain tumour &91.22 & 89.85 & 90.72& 90.28 &  93.98  & 93.54 & 92.94 & 93.24 & 93.66 & 93.63 & 91.89 & 92.76  \\

digital knee x-ray & 61.31 & 60.66 & 59.57 & 60.11 & 70.83   & 72.71 & 68.47 & 70.53 & 71.43  & 72.55 & 70.98 & 71.76 \\

Mini-DDSM & 66.88 & 67.42 & 67.46 & 67.44  & 91.05   & 91.09 & 91.30 & 91.20 & 91.94  & 91.96 & 92.16 & 92.06 \\

\hline 
\end{tabular}
}
\end{center} 
\end{table*} 


\begin{table*}[htbp]  
\begin{center} 
\caption{The classification performance of other training strategies using the DenseNet-121 baseline.}
\label{Ablation_DenseNet} 
\small
\tabcolsep=0.10cm
\resizebox{12.0cm}{!}{
\renewcommand{\arraystretch}{1.1}
\begin{tabular}{c |  c  c  c c | c c c c | c c c c } 
\hline 

Dataset  & \multicolumn{4}{c|}{Traditional transfer learning } & \multicolumn{4}{c|}{\emph{CLOG-CD}} & \multicolumn{4}{c}{ \emph{CURVETE}(WO/CL, W/SD)}  \\
&  \multicolumn{4}{c|}{}  & \multicolumn{4}{c|}{} \\

& ACC & PR & RE & F1 & ACC & PR & RE & F1 & ACC & PR & RE & F1 \\
 &($\%$) & ($\%$) & ($\%$) & ($\%$) &  ($\%$) & ($\%$) & ($\%$) & ($\%$) & ($\%$) & ($\%$) & ($\%$) & ($\%$)  \\

\hline 
brain tumour & 89.59 & 88.35 & 88.15 & 88.25 & 91.87 & 90.45  & 92.12 & 91.28 & 93.33 & 92.23 & 92.33& 92.28  \\

digital knee x-ray & 69.05 & 72.63 & 68.00 & 70.24 &67.26 &  67.15 & 64.14 & 65.61 & 73.21  & 71.13 & 70.43 & 70.78   \\

Mini-DDSM & 66.75 & 67.41 & 67.19 & 67.30 & 84.65 & 84.90 & 84.86& 84.88 & 86.32 & 86.60 & 86.57 & 86.58  \\

\hline 
\end{tabular}
}
\end{center} 
\end{table*} 

In addition, we conducted a Wilcoxon signed rank test at 0.05 \cite{wilcoxon1992individual} to evaluate the significance of \emph{CURVETE}'s performance compared to traditional transfer learning, \emph{CLOG-CD}, and \emph{CURVETE}(WO/CL, W/SD) models. For the brain tumour dataset (ResNet-50), CURVETE achieved statistically significant improvements with \emph{p}-values of 0.038, 0.001, and 0.0039, respectively. On the digital knee x-ray dataset, the corresponding \emph{p}-values were 0.0025, 0.00021, and 0.0091. For Mini-DDSM, they were 0.0010, 0.0053, and 0.0010. 

A similar trend was observed using DenseNet-121 across all three datasets. The \emph{p}-values of \emph{CURVETE} were 0.0217, 0.0014, and 0.0037 for the brain tumour dataset. On the digital knee x-ray, \emph{CURVETE} achieved 0.0079, 0.0032, and 0.0081 against traditional transfer learning, \emph{CLOG-CD}, and \emph{CURVETE}(WO/CL, W/SD), respectively. On the Mini-DDSM, the values were 0.0029, 0.00173, and 0.0035 for \emph{CURVETE} against other models. As you can see, all \emph{p}-values are below the 0.05 threshold, confirming that the improvements achieved by \emph{CURVETE} are statistically significant compared to other models.

\subsection{Comparison with State-of-the-art Methods}
We compared our proposed method with other state-of-the-art SSL approaches. This comparison is given in Table \ref{comparison}. First, we compared \emph{CURVETE} with our previous work \emph{4S-DT}, which used self-supervised sample decomposition to improve the classification performance of chest x-ray images. In addition, \cite{lim2023scl} introduced SCL, combining rotation degree prediction with contrastive learning to minimise the distance between training images and their augmented versions. In \cite{preciado2022self}, the authors presented RotNet with different pre-trained networks to learn image representations by predicting the rotation angles applied to input images. Similarly, \cite{breiki2021self} explored three different methods for applying SSL. We experimented with SimCLR and SRGAN methods. SRGAN upsamples low-resolution images to create high-resolution ones for fine-grained classification, while SimCLR uses contrastive loss to generate two augmented versions of each image and ensures that their representations are closer in feature space while remaining distinct from representations of other images in the batch. As shown in Table \ref{comparison}, \emph{CURVETE} outperforms all other methods on all three datasets.

\begin{table*}[htbp]
\begin{center} 
\caption{Comparison with other state-of-the-art methods.}
\label{comparison}
\resizebox{9.0 cm}{!}{
\begin{tabular}{c|c|c|c|c} 
\hline
Reference  & Method & \multicolumn{3}{c}{Acc (\%)} \\
& & brain tumour & digital knee x-ray & Mini-DDSM\\
\hline
\cite{abbas20214s} & 4S-DT  & 91.38 & 68.45 & 71.36 \\ 

\cite{lim2023scl} & SCL & 94.79 & 46.88 & 73.83\\

\cite{preciado2022self} & RotNet (DenseNet-121)    & 95.28 &  45.24 & 36.32  \\

\cite{preciado2022self} & RotNet (ResNet20)  & 79.35 & 58.93 & 60.36 \\

\cite{breiki2021self}  & SSL (SimCLR)  & 64.07 & 33.93 & 48.34   \\ 

 \cite{breiki2021self}  & SSL (SRGAN) & 43.41 & 65.03 & 72.90  \\  

\textbf{Ours} & \emph{CURVETE} (ResNet-50) &\textbf{96.60} &  75.60  &  \textbf{93.35} \\ 

\textbf{Ours} & \emph{CURVETE} (DenseNet-121) & 95.77 & \textbf{80.36} &  93.22  \\ 
\hline
\end{tabular}
}
\end{center} 
\end{table*}

\section{Discussion and Conclusion}
\label{discussion}
Training a convolutional neural network using a transfer learning strategy has demonstrated remarkable success in a variety of applications and provided a practical solution when dealing with limited labelled datasets. However, irregularity in distribution between classes remains a significant concern in machine and deep learning algorithms, especially in medical image classification tasks, where the model performance can be affected by certain classes, leading to inaccurate predictions. In this work, we proposed a progressive self-supervised training based on the curriculum learning strategy, called \emph{CURVETE}, to address this challenging problem. \emph{CURVETE} utilises the anti-CL strategy based on descending-ascending order for training the pretext model and the downstream task. The method was evaluated on three different medical image datasets: brain tumour, digital knee x-ray, and Mini-DDSM. In addition, we investigated the performance of (\emph{CURVETE}) compared to different training processes: traditional transfer learning technique, \emph{CLOG-CD}, and \emph{CURVETE}(WO/CL, W/SD) models.
The experimental results demonstrated the ability of the \emph{CURVETE} model to enhance representation information and extract high-level features in the pretext training task, guided by the anti-CL strategy, leading to better convergence in a new classification task. For the brain tumour dataset, \emph{CURVETE} has achieved an overall ACC of 96.60\%, an increase of 2.6\% over the classification performance without using the CL strategy in the pretext task. For the digital knee x-ray image, \emph{CURVETE} achieved an accuracy of 80.36\% using DenseNet-121, outperforming \emph{CLOG-CD} by more than 9\% and \emph{CURVETE}(WO/CL, W/SD) by more than 6\%. Likewise, for the Mini-DDSM dataset, the overall accuracy of \emph{CURVETE} was 93.35\%, higher than other training strategies. Moreover, \emph{CURVETE} with baseline DenseNet-121 shows a significant improvement in brain tumour and Mini-DDSM datasets with 95.77\% and 93.22\%, respectively, and surpassed the other training models in digital knee x-ray images with 80.36\%.
%
%
%
\bibliographystyle{splncs04}
\bibliography{references}

\end{document}